\documentclass[Times,12pt,oneside,openany,print,index]{report}
\pdfoutput=1
\usepackage[utf8]{inputenc}

\usepackage[T1]{fontenc}

\usepackage[a4paper,width=150mm,top=25mm,bottom=25mm]{geometry}
\usepackage[english]{babel}
\usepackage[utf8]{inputenc}
\usepackage{csquotes} 
\usepackage{amsmath} 
\usepackage{subcaption}
\usepackage{graphicx} 
\graphicspath{/images} 
\usepackage{caption} 
\usepackage{array} 

\usepackage[nottoc]{tocbibind} 

\usepackage[normalem]{ulem} 

\usepackage{hyperref} 
\hypersetup{
    colorlinks=true,
    linkcolor=black,
    filecolor=magenta,      
    urlcolor=black,
    citecolor=black,
}
\usepackage{nomencl} 
\let\cleardoublepage=\clearpage 

\begin{document}
\thispagestyle{empty} 
\begin{titlepage}
\renewcommand*{\thepage}{Title} 

    \begin{center} 
        \vspace*{3cm} 
        
        {\fontsize{16pt}{22pt}\selectfont{Automatic Question \& Answer Generation Using Generative Large Language Model (LLM)
}
        } 
        
        \vspace{1.5cm}
        
        \text{by}
        
        \vspace{0.5cm}
        
        	A.S.M Mehedi Hasan\\
	        20101128\\
	        Md. Alvee Ehsan\\
	        20101123\\
	        Kefaya Benta Shahnoor\\
	        20101115\\
	        Syeda Sumaiya Tasneem\\
	        20101346

        \vspace{1.5cm}
        
        	A thesis submitted to the Department of Computer Science and Engineering\\
            in partial fulfillment of the requirements for the degree of\\
            B.Sc. in Computer Science and Engineering

        \vspace{2.5cm}
        
    		Department of Computer Science and Engineering\\
                School of Data and Sciences\\
            Brac University\\
            January 2024
        
        \vspace{3cm}
        
    		\copyright\ 2024. Brac University\\
            All rights reserved.
    
    \end{center}

\end{titlepage} 
\cleardoublepage

\pagenumbering{roman} 




\phantomsection
\addcontentsline{toc}{chapter}{Abstract}
\section*{Abstract}
In the realm of education, student evaluation holds equal significance as imparting knowledge. To be evaluated, students usually need to go through text-based academic assessment methods. Instructors need to make diverse sets of questions that need to be fair for all students to prove their adequacy over a particular topic. This can prove to be quite challenging as they may need to manually go through several different lecture materials. Our objective is to make this whole process much easier by implementing Automatic Question Answer Generation (AQAG), using fine-tuned generative LLM. For tailoring the instructor’s preferred question style (MCQ, conceptual, or factual questions), prompt Engineering (PE) is being utilized. In this research, we propose to leverage unsupervised learning methods in NLP, primarily focusing on the English language. This approach empowers the base Meta-Llama 2-7B model to integrate RACE dataset as training data for the fine-tuning process. Creating a customized model that will offer efficient solutions for educators, instructors, and individuals engaged in text-based evaluations. A reliable and efficient tool for generating questions and answers can free up valuable time and resources, thus streamlining their evaluation processes.

\vspace{1cm}
\textbf{Keywords:} Automatic Question Answer Generation(AQAG); Natural Language Processing(NLP); Machine Learning(ML); LLM(Large Language Model); Prompt Engineering (PE); ReAding Comprehension dataset from Examinations(RACE) 
\pagebreak


\phantomsection
\addcontentsline{toc}{chapter}{Acknowledgment}
\section*{Acknowledgement}
In the beginning, sincere thankfulness to almighty Allah for his seamless guidance while we completed this research.
Secondly, we extend our profound gratitude to Dr. Farig Yousuf Sadeque, our supervisor, for his excellent assistance, valuable insights and direction during our work. We thank Md. Mustakin Alam, our co-supervisor, for his guidance and constructive feedback.
Finally, we are thankful to our loving parents for their constant support and prayers which enabled us to reach this point in our journey.

\renewcommand{\contentsname}{Table of Contents} 
\cleardoublepage
\phantomsection
\addcontentsline{toc}{chapter}{Table of Contents} 
\tableofcontents 

\listoffigures 
\listoftables

\makenomenclature
\renewcommand{\nomname}{Nomenclature}

\renewcommand{\nompreamble}{The next list describes several symbols that will be later used within the body of the document}
\printnomenclature
\phantomsection
\addcontentsline{toc}{chapter}{Nomenclatures}
\nomenclature{\(AQAG\)}{Automatic Question Answer Generation}
\nomenclature{\(NLP\)}{Natural Language Processing}
\nomenclature{\(AI\)}{Artificial Intelligence}
\nomenclature{\(ML\)}{Machine Learning}
\nomenclature{\(MCQ\)}{Multiple Choice Question}
\nomenclature{\(DUC\)}{Document Understanding Conference}
\nomenclature{\(PSVM\)}{Probabilistic Support Vector Machine}
\nomenclature{\(NBC\)}{Naive Bayes Classifier}
\nomenclature{\(AE\)}{Auto-encoders}
\nomenclature{\(AWR\)}{Average Weighted Recall}
\nomenclature{\(KNN\)}{K-Nearest Neighbors}
\nomenclature{\(SMO\)}{Sequential Minimal Optimization}
\nomenclature{\(ROUGE\)}{Recall-Oriented Understudy for Gisting Evaluation}
\nomenclature{\(CNN\)}{Convolutional Neural Network}
\nomenclature{\(RNN\)}{Recurrent Neural Network}
\nomenclature{\(BERT\)}{Bidirectional Encoder Representations from Transformers}
\nomenclature{\(SQuAD\)}{Stanford Question Answering Dataset}
\nomenclature{\(LLM\)}{Large Language Model}
\nomenclature{\(LoRA\)}{Low Rank Adaptationn}
\nomenclature{\(LoRA\)}{Low Rank Adaptationn}
\nomenclature{\(JSON\)}{JavaScript Object Notation}
\nomenclature{\(RACE\)}{ReAding Comprehension dataset from Examinations}
\nomenclature{\(XAI\)}{Explainable AI}
\nomenclature{\(TF-IDF\)}{Term Frequency-Inverse Document Frequency}
\cleardoublepage

\pagenumbering{arabic} 

\chapter{Introduction}
\section{Overview} 
A key function of the natural language process (NLP) is to produce sets of meaningful questions and answers from a given text. On the other hand, large language models (LLMs) have revolutionized the area of natural language processing (NLP) in recent years. Certain models have shown unparalleled ability to comprehend and produce writing that is human-like. The automatic generation of question and answer pairs is an intriguing application that has arisen from this innovative technological leap, and it represents an important milestone in the progress of machine learning. The fundamental working principle of LLMs is training on a variety of datasets which allows them to catch up on subtleties in language, context, and domain-specific information. Consequently, these models demonstrate an impressive ability to comprehend the nuances of many subjects, rendering them appropriate choices for the development of systems that generate questions and answers in a variety of fields, including science, technology, literature, and other fields. Exploring automatic question-answer generation is motivated by the growing need for intelligent systems that can understand and react to natural language questions. The scalability and adaptability of traditional techniques to question answering were limited since they frequently depended on rule-based systems. On the other hand, LLMs use enormous volumes of pre-existing linguistic data to understand intricate patterns and semantic linkages on their own. By fine-tuning these models for certain tasks, like answering questions, their capacity to produce precise and well-reasoned answers is improved. This allows them to produce queries and responses that are appropriate for the given context. But even with these amazing improvements, there are still difficulties when it comes to incorporating Large Language Models into Q\&A systems. The potential to provide inaccurate or misleading information, biases in training data, and ethical considerations are important issues that need to be carefully considered. However, automated question-answer generating has a wide range of significant applications. This question and answer creation technique will help users in many ways. For example, it will help to improve understanding of the topic by answering questions and promoting critical thinking. This technique is mainly useful for teachers because they need to generate several sets of questions from various texts within a short period of time, sometimes this may be a big challenge for them. In this paper , we propose  a system where diverse types of questions can be constructed which can result in creating more comprehensive assessments in an efficient way compared to manual approach. At first, our system will fine tune the large language model with our dataset using prompt engineering for the question answer generation process. This attempts to provide more relevant outputs.Then, it will generate multiple questions in different formats and their corresponding answers. Our work will focus on the English language by the help of NLP and ML.
\section{Problem Statement}
In the realms of education, evaluating a student’s knowledge regarding any topic requires a process of question-answer phase. This whole approach creates a real burden to the educators or instructors as they need to formulate questions: subjective and objective. While preparing the questions, they have to be careful about the question patterns to ensure that meaningful, information-based questions are produced. The responsibility is not bound to the question generation, they need to evaluate the students’ responses as well. This whole manual process takes immense time and effort for an educator, sometimes the diversity of questions might be hampered. It could be challenging for instructors to maintain the dynamic and diversity of questions that properly capture a student's understanding.
\\
From the learners’ perspective, they sometimes are unable to get their assessments feedback from educators in time as sometimes educators can’t provide the feedback due to business. After studying a particular topic, a student might need to evaluate himself so that he/she can be well known about his/her preparation before any important formal assessment task. By this, students can find out their lackings about any topic so that they can give more time on those particulars. Moreover, they will have knowledge about the significance of the article by analyzing the comprehensive text-based questions, helping them to acquire more knowledge and enrich their writing skills. 
\\
Beyond the educational field, “Automatic Question and Answer Generation” model can make a significant contribution in the corporate or professional world. For hiring employees, it is a great challenge for the company to choose the best suited candidate for any particular position among thousands of applicants. The selection criteria must be arranged in such a way that there is no chance of questioning biasness. Following these concerns, our proposed model can be a great solution for arranging an assessment for the candidates based on some key words for that following position. Thus, it lessens the dependence on the information of CV and prioritizes the required skills of the role. Thereby, all candidates can get equal opportunity to hold eligibility for the job. 
\\
To summarize, our proposed model “Automatic Question and Answer Generation'' system will add a new dimension to this new era, ensuring a dynamic and significant learning area. This research will have an overview of developing and implementing the model for the purposes that we have discussed above.

\newpage
\section{Research Objective}
In this research, we develop a contextually suitable question-answer generator from a given document. Our strategy for summarization and question generation make use of unsupervised learning techniques in the language processing domain. Specifically, the main contributions of this research are: 

\begin{itemize}
    \item The model sought to develop an “Automatic Question Answer Generator” using Large Language Model(LLM).
    \item Our model will be able to generate meaningful questions and their corresponding answers from a given text.
    \item The education sector will take on a new dimension using this model as the faculties will be able to automatically generate their preferred questions and the students will be able to practice and evaluate themselves on their own. 
    \item The model will save a lot of time and hard-work for the user. 
    \item After completion, we aim to release the model and publish our work.
\end{itemize}

\newpage
\section{Research Contribution}
In this research, we develop a contextually suitable question-answer generator from a given document. Our strategy for question generation makes use of unsupervised learning techniques in the language processing domain. Specifically, the main contributions of this research are: 

\begin{itemize}
    \item We developed a reliable tool “Automatic Question Answer Generator” for extracting relevant information from context articles and automatically generating insightful questions and the corresponding answers.

     \item The developed model shows similar results to the 4-bit quantized base Llama-2 model, with the addition of our fine-tuning parameters. 
    \item As Meta’s Llama-2 is quite new, with an open-source release of less than 1 year, we strived to contribute to research literature to expand upon the scholarly understanding of this particular LLM’s capabilities.
 
\end{itemize}

\section{Thesis Structure}
The thesis is organized into 6 chapters. The first chapter "Introduction”, states an outline of the research statement, the objectives of the research and the research contributions. The second chapter is titled “Related Work” where some literature focusing on the research problem is reviewed. Chapter three, titled "Dataset, Data Analysis,and Data Preprocessing" states an overview of our dataset and the exploratory data analysis. Chapter four “Methodology, Architecture and Model Specification” outlines how the research problem was examined and addressed along with the workflow of the model. Chapter five “Result Analysis”, is about the results and performance of our model. Finally, in Chapter 6 "Conclusion", the results and conclusions of our work on the entire research problem are summed up along with the problems we faced during the research process and some approaches we tried to overcome those problems are mentioned.

\chapter{Related Work}
\section{Extractive Text Summarization}
To realize our scope in the field of extractive text summarization, we have researched some scholarly articles. For selecting, we have tried our best to choose those articles which have a decent number of citations, published in recent years. \\
The paper \cite{r1} proposes a learning-based method for combining different sentence features. They suggest that combining the sentence features is beneficial because each one of them makes a distinct contribution. Determining integrated sentence attributes for extractive summarization is what they research. They use a supervised learning framework to estimate the weights of various characteristics in order to evaluate how probable it is that a sentence will be significant. To determine which sentences are crucial for summarization, they use supervised learning to test the efficacy of various sentence attributes. Sentence feature vectors are analyzed, and then a supervised learning classifier is used. Candidate sentences are re-ranked in particular since final summaries' lengths are fixed. The final summaries are generated by extracting the strongest sentences. Combining characteristics considerably improves summarization performance, according to experiments. The tests have made use of DUC 2001 dataset. It has 308 total papers in 30 clusters of pertinent documents. Each cluster includes model summaries written by NIST assessors and addresses a particular subject (such as a hurricane).  In supervised learning and semi-supervised learning studies, the structure for training and testing is the same. To assess the summarization performance, ROUGE (Lin and Hovy, 2003), an automated assessment software, is used. Based on the overlap, it contrasts model summaries with summaries produced by machines. They create a training strategy to train several classifiers using the same feature space. The combination of surface, content, and relevance aspects is implemented using PSVM and NBC. According to experiments, features that combine surface, content, and relevance function the best and compared to supervised learning, the semi-supervised learning strategy retains equivalent performance while saving half of the labeling cost. 
\\\\
Mahmood Yousefi-Azar and Len Hamey \cite{YOUSEFIAZAR201793} authored a paper focusing on an unsupervised model featuring its deep learning capabilities. The paper highlights query-based single-page documents using an automatic metric including Auto-encoders [AE] for text summarization. Auto-encoders are methods that learn to create efficient data representation. Taking input data the encoder model compresses it into a smaller, more efficient state. With this smaller state of representation, the decoder tries to recreate the input state as closely as it possibly can. Through this process [AE] learns efficient data outputs. Word representation is a key factor of the proposed model. As other systems rely on sparse representations of data, it causes problems in the training process where there enough data input is not observed. This subsequently increases the problem where only the sub-parts of the text are used as training data. The proposed model structure combats this by representing only distinct local word structures as input and by adding random noise in the vector representation. The effect of adding random noise into the input vector is also used multiple times with different added noises. [AE] has been used in unsupervised text summarization models before but only as a word filter. The application of AE’s, and the [ENAE] method in word ranking is the primary focus of this paper. Consequently, it becomes a stochastic model rather than a deterministic feed-forward network composition. Distinct models extracted summaries of different lengths. Local vocabulary enhances recall by an average of 11.2\% when [AE] is used. It causes no adverse effects when the encoder maps sentences of different documents to different semantic spaces. BC3 datasets, For comparison analysis, average weighted Recall (AWR) was used as a metric for supervised and unsupervised analysis.The authors draw the conclusion that their novel strategy is on par with the best supervised strategies. With hopes that reduced computational cost makes the new model more suitable, where semi-supervised learning implementation can be explored in the future.
\\\\
Another paper by Devi Krishnan, Preethi Bharathy, Anagha and Manju Venugopalan \cite{9065651} states extractive summarization from the text based on some robust features which is a supervised approach. The whole model works as a flow of some steps. At first, the training data needs to be preprocessed for feature extraction. In the preprocess steps, tokenization, part of speech tagging, stemming and stop-word removal have been followed. To resolve the class imbalance, it has mentioned a ‘Sampling’ method which can reduce the majority classes or increase the minority classes according to necessity. To train the extractor, the model has used some significant algorithms like Naïve Bayes, KNN, Random Forest, SMO, J48, Bagging. To measure the accuracy of the trained model in extractive summarization, the used data set is “BBC news article” from 2004 to 2005 containing 2225 documents divided in 5 domains. The proposed model summary has been compared to the dataset summary and on ROUGE scale it has given average scores beyond 0.4 which is far better than other trending research. It has also stated that ROUGE-1 has provided the best result as it works based on unigram overlapping.
\\\\
Akanksha Joshi, E. Fidalgo, E. Alegre and Laura Fernández-Robles has introduced a new innovative technique SummCoder for creating extractive summarization \cite{JOSHI2019200}. This technique requires three metrics developed by the researcher of this paper to select sentences for generating the summary. First metrice is sentence content relevance and to determine this they use a deep auto-encoder network.Second metrice is sentence novelty is used for measure the similarity between sentences and third one is sentence positive relevance, this feature is designed by hand  and it gives more importance to the initial sentences using weight calculation method which generated through the length of the paper or document dynamically. To test their method more accurately they introduced a new dataset TIDSumm. In their research they used 5 datasets. For training and validation their approach they use CNN and DailyNews dataset and for testing purposes they use DUC2002 dataset which contains news, Blog Summarization dataset containing blogs and TIDSUM dataset which they made, containing illegal activities.Using DUC2002 dataset, heir method gained a more  brilliant score in Rouge than other algorithms for extractive summarization on  single documents. Their model gained 51.7 in ROUGE-1, then 27.5 in ROUGE-2, 28.5 in ROUGE-SU4 and lastly 44.6 in ROUGE-L. Their model shows better performance than other text summarizers. Though it does not score better in ROUGE-1 than the CoRank approach. By using their self build dataset TIDSumm they gain scores in ROUGE-1 is 58.8, ROUGE-2 is 48.9, ROUGE-SU4 is 45.9 and ROUGE-L is 49.3. By seeing this score they are clear that their approach is better than compared to other baseline algorithms.They also observed when they use Blog Summarization dataset their approach SummCoder gets the highest score in ROUGE score compared to the other baseline approaches.
\\\\
Jiacheng Xu , Zhe Gan, Yu Cheng, Jingjing Liu \cite{xu-etal-2020-discourse} proposed extractive models which are sentence based sometimes produce ineffective and redundant sentences in the summary and also Bidirectional Encoder Representations from Transformers(BERT) a language model fails to capture wide dependencies properly because it is trained on pairs of sentences but not for the full size documents. This DISCOBERT model works with the smaller chunks of a sentence which they call discours unit and from this it has the ability to minimize redundant details which are present in the sentences. They have three contributions in this research.This model generates two types of discourse related graph which shows the connection and relatedness among sentences. And using this information this model can generate summaries that capture the sense of the original documents. According to their research this model is better than the previously built best performing models when it comes to generate the qualityful summaries. They compare their model with other BERT based models and it gives better performance on generating summaries. New York Times (NYT),CNN and DailyMail (CNNDM) these two datasets they used in their research to test their model. From See et al. (2017)  they use 5026 script to extract summaries from raw data and they use Stanford CoreNLP for detecting the boundaries of the sentence, then tokenizing the sentence and parsing the syntactic structure.when they use CNNDM dataset the result they get from their DISCOBERT model that beats the other state of the art BERT model by gaining higher scores differences 0.52 on R-1, 0.51 on R-2 and 1.04 on R-L on the F1 evaluation metric. Using the NYT dataset they get the result of a significant gain of 1.30 on R-1, 1.29 on R2 and 1.82 on R-L when it compares with the BERT baseline. 
\\\\
The paper by Xingxing Zhang, Mirella Lapata, Furu Wei, and Ming Zhou \cite{zhang-etal-2018-neural} proposes an extractive model for latent variables and consider the labels of sentences as binary latent variables. An extractive model is used to predict the latent variables, and gold summaries are directly responsible for the training loss. The authors claim that existing extractive summarization models are minimal as they rely on sentence-level labels. The model may be thought of as a reinforcement learning model. It includes a fundamental linear regression model to lessen the variance of gradients. The authors initiate the latent model using a pre-trained extractive model to prevent random label sequences while sampling. The paper starts by introducing the neural extractive summarization model, which serves as the foundation for the latent model. Then it next goes through the sentence compression model that the latent model uses. The authors tested their model on the CNN/Dailymail dataset and the results revealed that it outperformed an effective extractive baseline trained on rule-based labels.  By a large margin, the extractive model performs better than LEAD3. Additionally, EXTRACT performs better than previously released extractive models like SummaRuNNer, EXTRACT-CNN, and REFRESH. The suggested model can indeed outperform a robust extractive model, according to experimental findings. 
\\\\
The paper by Ramesh Nallapati, Feifei Zhai and Bowen Zhou \cite{nallapati2016summarunner} provides an innovative approach for extractive summarizing that makes use of a recurrent neural network RNN. The authors believe that this method is more interpretable and can produce greater results than earlier approaches. The model is capable of extracting sentences that are both prominent and informative. It was trained on a dataset of human-generated summaries. The proposed method consists of two key phases. The RNN is initially taught to predict a probability distribution across the document's phrases. Then a greedy algorithm is applied in order to choose the sentences with the greatest probability ratings. The authors tested SummaRuNNer on several different datasets and discovered that it performed better than earlier approaches. Additionally, they discovered that SummaRuNNer was easier to understand than earlier techniques since it could recognize the sentences that it thought were crucial. According to the paper's findings, SummaRuNNer, a recurrent neural network (RNN) based sequence model for extractive document summarization, delivers state-of-the-art performance on the CNN/Daily Mail dataset. The study provides two more contributions in addition to the suggested method. First, the authors suggest a novel technique for abstractive training of extractive models. Sentence-level extractive labels are not necessary with this model since training can be done using only human-generated reference summaries. Additionally, they intend to build a joint extractive-abstract model in which the predictions of their extractive component generate stochastic intermediate units that the abstractive component consumes. The authors also give a thorough assessment of extractive summarization techniques.
\\\\
Shashi Narayan Shay, B. Cohen, Mirella Lapatathis\cite{narayan-etal-2018-ranking} has introduced a unique training algorithm which uses reinforcement learning objectives to globally maximize the ROUGE evaluation metric. They use CNN and DailyMail dataset and on this dataset, they use this algorithm to build a neural summarization model. They demonstrate through trials that their model performs better than the extractive and abstractive systems currently in use, both when judged automatically and by the help of the humans. The researcher of this paper contended that in their study the cross-entropy is an inadequate approach for the extractive summarization.  This method of training the models makes them more likely to provide more wordy summaries that are the reason for long sentences and those summaries contain a lot of redundant information. For summary creation they rank sentences through a reinforcement learning objective and ROUGE evaluation metric to overcome the mentioned challenges. Their algorithm which they used for training has the ability to explore the grounds of summaries and make their model stronger to new information. For this reason, reinforcement learning helps the process of extractive summarization in two ways.firstly,the evaluation metric is optimized straight and secondly it makes their model a better version while the model selects the sentences. The researchers use the CNN and DailyMail news highlights datasets to assess their model and for training and testing they use standard splits of Hermann et al. (2015). According to the result their model REFRESH is more efficient than LEAD baseline. In human evaluation, participants answer 66.34\% of the question that is summarized through REFRESH and summary that are produced by LEAD and other systems,  participants can answer the question 36.33\% and 28.73\% respectively. 
\\\\
Another paper authored by Changjian Fang, Dejun Mu, Zhenghong Deng, and Zhiang Wu\cite{FANG2017189} proposed a new Co-Ranking method for sentence ranking. This model named CoRank explores the relationship between graph-based general algorithms e.g. PageRank with weighted biases in words. Highlighting that general graph-based ranking methods do not consider sentence importances biases, weights of every word are considered equal. The authors correlated how the linkage between words in sentences deserves higher importance and scores in the Automatic Extractive Text Summarization [AETS] matrices. Sentences containing high-weighted words and the number of appearances in a sentence should get higher scores in the proposed ranking model. Extractive summarization methods operate by creating an intermediate node representation of the input text data and assigning ranking scores by adding biases. Unlike the more commonly used ranking algorithms, e.g. TextRank and TextRankExt, redundancy elimination and adding the dual criteria of word-sentence relationship justifies the effectiveness. The CoRank model, working under the bag-of-words [BOW] structure, includes the word-sentence score feedback system to improve its weighted graphs, giving importance to rankings of distinct words. A structure of redundancy elimination is also proposed to justify the effectiveness of the proposed methods in real-world applications, resulting in a sub-technique CoRank+. Utilizing the News and DUC02 datasets, and the analysis of external validation measures, the efficacy of the CoRank method in some cases and the CoRank+ method in all cases outperforms other relevant ranking methods. An unsupervised learning approach to the [AETS] structure is fundamental to this system as the proposed CoRank model can be considered as an addition to the unsupervised graph-based ranking. Although currently focusing on single-page documents with enhancements in word-sentence rankings, the authors conclude with a vision to improve their model to work with multi-page text collection summarization.
\\\\
The paper, \cite{dong-etal-2018-banditsum} authored by Yue Dong, Yikang Shen proposes a new model of single-document extractive summarization by training neural networks . The model is referred as BANDITSUM which considers extractive summarizing as a contextual bandit issue in which the model is given a context to determine a some sentences to generate summary. The model is trained to choose sentence sequences that optimize ROUGE score using a policy gradient reinforcement learning technique. The sequential binary labeling setup is eliminated in this technique. This action drastically decreases the amount of space that must be investigated, eliminates the need for supervised pre-training, and stops early phrases from being systematically given preference over later ones. The authors of the paper evaluate their approach on the CNN/Daily Mail dataset. For training, validating, and testing, they follow Hermann et al.'s (2015) conventional split; however, they do not apply anonymization on the three corpora. The outcomes demonstrate that BanditSum outperforms other similar approaches in terms of ROUGE scores. BANDITSUM outperforms SummaRuNNer, the cutting-edge maximum likelihood-based extractive summarizer, as well as two RL-based techniques, by a large margin. According to empirical findings, their methodology converges substantially faster than competing methods while outperforming or on the same level of some decent extractive summarization models's performance. 
\\\\
The research paper, written by Hans Christian, Mikhael Pramodana Agus and Derwin Suhartono introduced TF-IDF algorithm which is a new technique for extractive text summarization. It is a statistical measure that identifies the importance of a word in a paragraph and selects sentences containing those important words and then extracts those sentences. It helps  to read any document efficiently and shortly using this program. As, this program summarizes the document. This program can summarize more than one document while other summarizing tools can summarize one document only. The researchers of the paper  also focus on the accuracy of the summary. In this research they used only descriptive text. Throughout the research they show how they formed an extractive summary using the TF-IDF algorithm. They found that the algorithm is able to produce an accurate and informative summary of any documents and this summary is easier and shorter to read than the main articles. They run the program several times each time using different combinations of documents and compression rates. In their experiment, they use 6 different documents. They use the same documents in two different online summarizers beside their program  to compare the comparison rate. The results are, this summarizer is able to generate summaries that were 67\% accurate. The summary which is produced from this program is more accurate than the summary which is produced using two other online summarizers. According to them, this is very powerful to calculate the importance of a word in documents which helps to determine which sentence should be in the summary. 
\\\\
Authors M. Bidoki, M. Fakhrahmad and M.R. Moosavi \cite{Bidoki_Fakhrahmad_Moosavi_2020} explained the purpose of multi-objective optimization, through the application of harmony search and extractive multi-document summarization. Text summarizer methods aim to create a compressed version of the input statement. Hierarchical structure based on Score based rankings initializes the sentence's importance. After pre-processing the system trains on the Doc2Vec model, which is an extension of the Word2vec model that is used to generate vectors for words. Doc2vec generates vectors for documents by training a neural network on a large corpus of documents. The model learns to predict the context of a word or document in the corpus. The resulting vectors can be used to calculate the similarity between documents or clusters. Word2Vec is used as the word embedding in the numeric semantic vector. A sentences mean vectors measure the similarity, words with more relation with the main topic may have higher semantic loads. Harmony search is implemented to produce the optimal summary-building process. Utilizing the strengths of meta-heuristic algorithms in combination with multiple summarizer systems, the proposed framework tends to outperform other cited summarizer systems.
\\\\
The paper by Jesus M. Sanchez-Gomez, Miguel A. Vega-Rodríguez , Carlos J. Pérez has introduced a new algorithm named Multi-Objective Artificial Bee Colony (MOABC) to execute their task. This algorithm is specially built for extractive text summarization from multiple documents. Three distinct bee categories exist in MOABC which allows for various searches for each of the bee categories .The prime goal of this research is to reduce the redundancy and  simultaneously optimize content coverage and treat both of them as separate objectives. For evaluating the performance of their model they use the DUC 2002 dataset. To calculate the efficiency and result they evaluated their models using ROUGE metrics.This metrics determines how closely a summary produced by a computer and summary produced by humans are comparable. After counting the number of overlapping units it compares which is better and provides results. For the evaluation they use ROUGE-2 and ROUGE-L. In the result part the ranges and CVs clearly demonstrate that the suggested model’s output is very strong. The findings of the model give them significant improvements by considering an average improvement is in best single objective ROUGE-2 31.09\% and ROUGE-L 8.43\% compared with the  multi objective result ROUGE-2 18.63\% and ROUGE-L 6.09\%. It also shows that when they repeat the execution of the algorithm , the suggested model generates higher quality ROUGE values.It results in more reliable outcomes in the relative dispersion between  620.63\% and 1333.95\% of reduction which is better 6 and 13 times. 
\\\\
The proposed model by Ming Zhong, Pengfei Liu, Yiran Chen, DanqingWang, Xipeng Qiuy, Xuanjing Huang \cite{zhong2020extractive} is “MATCHSUM” which is a novel framework, focusing on semantic text matching. Summarization based on sentence-level score has some drawbacks as it prioritize sentences according the system score and extract sentences accordingly, ignoring semantic of the whole summary. Reinforcement Learning is used by Narayan et al. (2018b) and Bae at al. (2019), focusing on summary level scoring. But, this approach has not been so effective due to some limitation like auto regression or non-auto regression. Therefore, the paper has provided a visual representation of summary level approaches which boost the performance regarding semantical similarity between the main text and the summary. Before working on this model, some candidate summaries have been chosen by pruning the irrelevant information from the document which it introduced as “content selection module”. Six vast datasets have been used here to measure the performance score. It has portrayed tremendous performance on mid-level length summary like CNN. Besides that, when we arrange sentences according to the sentence-level score from higher to lower, the learning phase of sentence extractor becomes more challenging as it has to handle with pearl-summary(sentence-level score is low). On CNN dataset, there are 18.9\% which not considered as those kind of summary by which using sentence extractor might have risk for losing significant data. Using the BERT base, the proposed model has illustrates a clear view that on ROUGE metrics, “MATCHSUM” model has scored far better than all other models. Changing the encoders to baseline BERT or RoBERTa also gives the best outcome than others.
\\\\
Erkan, G{\"u}nes and Radev \cite{Erkan_Radev_2004} have introduced another method which is basically graph based. This  paper looks into how lexical centrality, which evaluates centrality based on the lexical characteristics of sentences, can potentially be measured in multi-document summarization. Here, three new metrics for sentence centrality: Degree, LexRank with threshold, and continuous LexRank are modeled after the social network idea of prestige. Moreover, it has introduced graph representation of a document cluster in place of traditional heuristics for centrality. The centroid basically determines the construction of words having greater  scores than threshold. And the sentences from which more words match with the centroid are considered as central. Therefore, to measure the similarity between sentences, a model named “Bag-of-words” has been introduced where the similarity between two vectors are calculated. Here, vector represents each word that constructs the whole sentence. Using cosine similarity scores, cosine similarity matrix can be generated which is used to calculate sentence centrality. Here, it denotes each edge as cosinse similarity matrix which is considered as vote between two sentences which is denoted as nodes. In this case, it set a threshold score value to ignore the overall less similarity scores compare to others as it finds it less necessary to be included in summary. But, the drawbacks of it is that sometimes it may include unnecessary information if those kinds of similar sentences will be found. To resolve this issue, eigen centrality concept can be used. Therefore, the centrality of a particular node(u) is divided in between nodes that are adjacent to u. Moreover, using the Perron-Frobenius theorem, it is proved that the Markov Chain with the stochastic matrix X converges to a unique stationary distribution. Because of this converging property, it works like an iterative algorithm where in every iteration, it results in a new eigenvector by multiplying it to  .  Using these “Power method” LexRank score is calculated for undirected sentence similarity graphs. It may boost the performance if the cosine values can be directly used in the graph. For testing the whole model, the chosen data set is DUC 2003 and 2004 where in total 70 clusters of documents have been used. Total eight experiments have been run for each cetrality method and in every experiment the proposed methodology have achieved higher scores on ROUGE-1 metrics than the baselines, even than the centroid based extractive summarization. It also highlighted that the lower thresholds for degree and LexRank features provide the best outcome.
\\\\
Yang et al. (2019) looks at the potential of BERT for text summarization in his paper \cite{liu2019text}. A number of natural language processing tasks, including text categorization, question answering, and natural language inference, have been successfully completed by BERT. The authors propose two different models: an extractive model and an abstractive model using BERT. The authors test the performance of their models on three datasets: CNN/DailyMail, NYT, and XSum. While XSum is largely abstract, CNN/DailyMail and the New York Times are rather extractive. The authors observe that their abstractive model performs similarly to the state-of-the-art on XSum while their extractive model beats the state-of-the-art on CNN/DailyMail and NYT. The impact of various BERT hyperparameters on summarization performance is also investigated by the authors. On all three datasets, they discover that using a bigger BERT model (BERT-large) results in superior performance. Additionally, they discover that optimizing BERT for the summarization job results in even greater performance gains. Overall, the study "Text Summarization with Pretrained Encoders" significantly advances text summarization research. The authors demonstrate how BERT may be utilized to produce cutting-edge outcomes on challenges requiring both extractive and abstractive summarization. Their research conveys new possibilities for using pretrained language models to problems involving natural language processing.
\\
\section{Multi Document Summarization}
Another study by Yang Liu \cite{liu2019finetune}  has highlighted an updated variant of the BERT model (BERTSUM) for extractive summarization which can enhance the performance better than previous. To enable obtaining summaries within multiple sentences, the model changes the BERT's input sequence and embeddings. In that instance, several summarization layers have been piled on top of the BERT outputs. This BERT outputs basically indicates the sentence vectors(T1,T2…..Ti) which can be achieved by following token embeddings, interval segment embeddings and position embedding. After that, for each sentence, a predictive score  (i=1,2,3…) is calculated. In order to calculate this score, it has followed a sigmoid equation depending on different variety layers (simple classifier, Transformer, LSTM).
BERD with Transformer layer results,
\begin{equation}
\centering
\hat{Y}_i=\sigma(W_0 h_i^l + B_0) 
\end{equation}
In the paper, it has shown the ROUGE F1 scores for each category layers applying on two large datasets, CNN/DailyMail and NYT. From the test set results it is clearly seen that BERTSUM with Inter-sentence Transformer layers have given the best outcome which has tested on, scores 43.25 on ROUGE-1 metrics and 39.63 on ROUGE-L metrics. They have also compared these BERTSUM results with some previously proposed systems like LEAD, REFRESH, NEUSUM. Additionally, it has been evaluated on the aforementioned dataset and shows that several BERTSUM components, such as interval segment and trigram blocking, can produce better results and more informative summaries, respectively. Here, interval segment indicates the distinction of sentences and trigram blocking is used to reduce the redundancy which is more simple compared to other pre-modeled algorithms, but simpler and more feasible.
\\\\
On the other hand, the paper\cite{FERREIRA20135755} talked about extractive text summarization on the basis of selection of sentence rankings. Extraction of summarized data usually starts from data pre-processing through the creation of paragraphs, sentences, tokens, and nodes. This paper dives into various different scoring methods widely used in the realm of single-page document summarization. 15 different scoring methods are described and implemented on three seperate datasets in this paper. For quantitative analysis, ROGUE assessment metric was implemented. ROGUE is used to quantitatively assess the summaries produced by the various scoring techniques. This fully automated evaluator essentially compares the system-developed summaries to the corresponding gold summaries in terms of content similarity. This paper hopes to fill up the gap between summarization techniques and comprehensive assessments. Considering large portions of stop words, co-reference ambiguity and redundancy the system implements sentence scoring methods like  word scoring; e.g. ( word frequency. Word Frequency, TF/IDF, Lexical Similarity, and Sentence Length are describes as best proven techniques. 
\\
\section{Question-Answer Generation}
Paper authored by Miroslav Blšták and Viera Rozinajová\cite{8501891} talks about the framework of an automatic question generation [AQG] structure analysis using machine learning approaches. This paper focuses on generating factual questions from unstructured English language text collections. After preprocessing, the framework begins by obtaining the lexical, syntactic, and semantic role labeler information of the input data. This focuses on pattern-matching tools for natural language processing. The authors explore through a comparison between the combination of traditional linguistics-based pattern recognition and a more statistical approach using neural networks. After the publication of the SQuAD data set, utilizing neural network methods started to gain popularity in question-generation tasks. Performance metrics such as BLEU-3, BLEU-4 and METEOR were used for performance analysis. In the proposed framework data sets consist of the sentence–question pairs extracted from text, and generating feedback using the reinforcement learning approach to improve the system. Sentence processing represents the sentence as a series of sequence tokens, named Composite pattern [CP]. There is no statistically significant method of evaluating [AQG] systems. Human evaluation can be an option, but due to large data sets it's not possible for humans to evaluate the questions out of subset instances. Multi-level classifications and distance metrics were used in the machine learning approach of assigning new question sentences. 
\\\\
Using the extractive text summarization technique, A. Nwafor, Chidinma and E. Onyenwe and Ikechukwu\cite{A_Nwafor_2021} published a paper focusing on generating multiple choice question answers using NLP. The whole process is divided into three main cores. One is extractive summarization to extract the keywords of the selected chapter or article to extract. Before doing summarization, it has followed some preprocessing techniques before following the extractive text summarizer features. Moreover, irrelevant data needs to be erased from the text before word normalization. The word is basically the tokenized form of a sentence where each word denotes as single token. For filtering the stop-words, IF-IDF concept has been used to measure the weight of a particular sentence. Moreover, it includes stemming for boosting the precision score and recall score of MCQ generations. After the extraction has been done, it is necessary to find the context of each sentence which are semantic and meaningful and then analyze which question will be standard leveled. To analyze it, it is important for having a list of keywords which may be created by the teacher or who wants auto generating question answers from the text. Then the precision and recall score will indicate the accuracy of the model finding the exact keywords from the text. As a test case, the system has chosen five different materials on which it has portrayed recall score indicating the majority of the extracted words are in the listed keywords and precision stating more important words rather than those selected keywords. To make the question, extracted keywords having higher IF-IDF values will be mapped to the real position in the text and these can be easily replaced by dash \------. After that, we can choose the number of options for the question and each option can be filled with other keywords whereas remaining one option must be filled with the main keywords for that blank space. It will be reviewed by teacher as well for approving the question.

\chapter{Dataset, Data Analysis and Data Pre-processing}
\section{Description of the Data}

We have chosen a race training dataset which has 10,000 reading comprehension and for each comprehension it has four questions and each question has corresponding four multiple options to choose. So in this dataset there  are 40,000 questions. The data initially existed in JSON format but was transformed into a Microsoft Excel CSV file format named "race-train40\_FINAL.csv.". These passages and questions were taken from English tests given to students in middle and high schools in China. Models for automatic understanding can be trained and tested using this dataset.

In the data features, each passage is assigned a distinct identification number within this dataset. Furthermore, an article is a text, essentially a passage of content.On the other hand, questions are the list of strings .In this case there are two types one of them is interrogative sentences and the other one is placeholders represented by dash which are called fill in the blanks. Additionally, there is a list of options in the dataset and Each list has 4 candidate options to choose from. Lastly, answers are a list containing the golden label for every question.

\section{Data Pre-processing and EDA}
\begin{figure}
    \centering
    \includegraphics[width=0.8\linewidth, height=10cm]{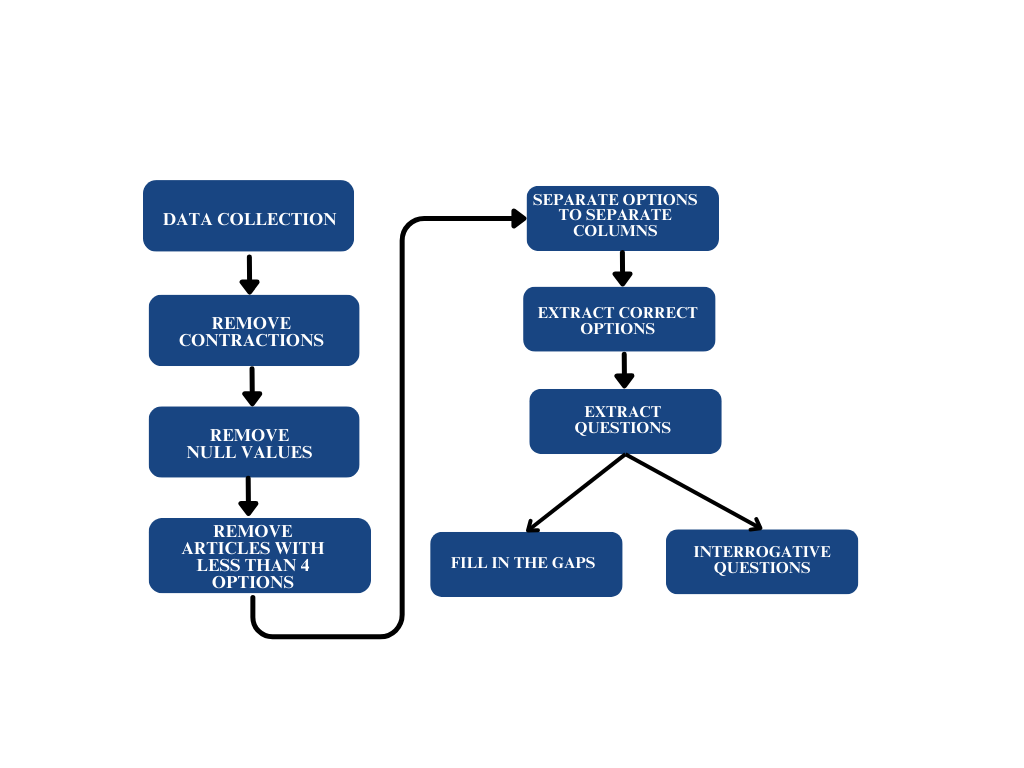}
    \caption{Preprocessing Workflow}
    \label{fig:enter-label}
\end{figure}
We reduced certain data that didn't include all the multiple-choice question (MCQ) options to ensure our model could train effectively. At the beginning of the pre-processing, we manipulate  the text by converting contractions to their full forms and handle the null values in each of the columns. A contraction is a shortened word or word group that is usually made by joining two words together and leaving out one or more letters. Although contractions frequently appear in spoken language and casual writing, they should be avoided in official or academic environments. It is important to remove contractions while pre-processing in the natural language processing context due to various reasons. It increases the unity and consistency of the text. Moreover, eliminating contractions makes tokenization easier since each one is handled as a single token which facilitates easier processing and analysis later on. On the other hand, in order to ensure the accuracy, dependability, and smooth operation of data analysis and modeling procedures, null values must be eliminated. After that, by using the pandas library we have created a column named “correct” where the correct answer option is written for particular questions. Then we counted the average length of an article in terms of the number of words. Not only this but also we have counted the average length of the text in articles, questions, and right answer choices, as well as the average length of all answer options for data analysis. 

\textbf{After pre-processing, we have decided to work with 33840 rows of input data for our training dataset.}
\vspace{5pt}

Average Article word count= 311.66, Average Question length= 10.500975234942963
\vspace{5pt}

\begin{figure}[h]
\centering 
\includegraphics[scale=0.7]{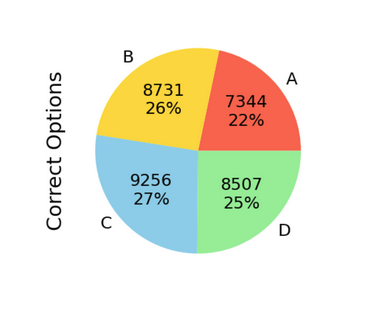}
\caption{Correct Options}
\label{fig:workflow1}
\end{figure}

\begin{figure}[h]
\centering 
\includegraphics[scale=0.45]{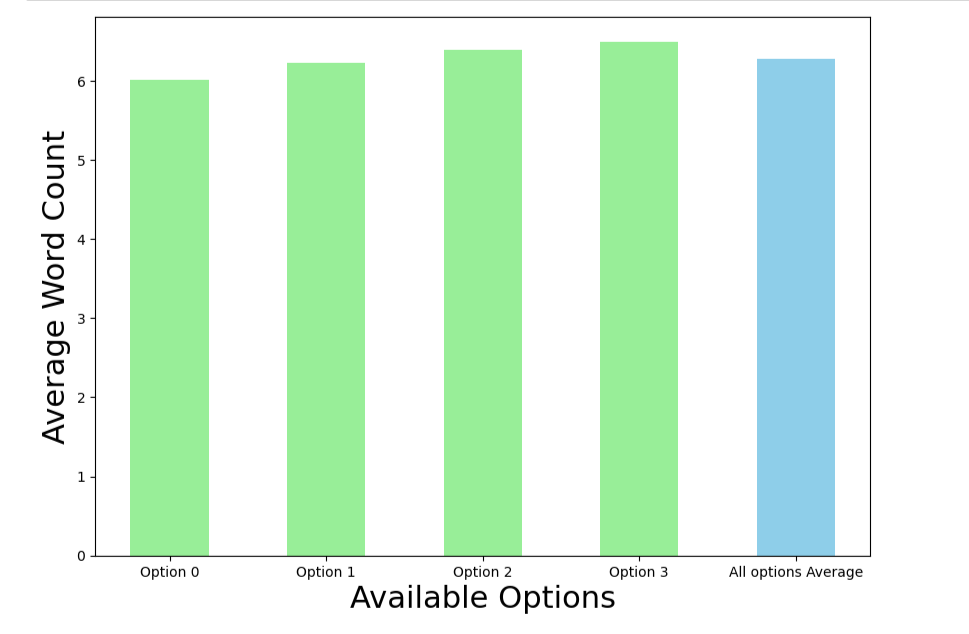}
\caption{Average length of option texts}
\label{fig:workflow2}
\end{figure}

The Fig. \ref{fig:workflow1} illustrates the distribution of a whole correct option into its constituent parts or categories visualise the options which occurred more frequently. Among the 4 options, we observe that chances are almost evenly split; Option A appears as the correct answer 22\% of the time, Option B appears as the correct answer 26\% of the time, Option C appears as the correct answer 27\% of the time, Option D appears as the correct answer 25\% of the time. The same thing is shown using a bar chart \ref{fig:workflow2}.

\begin{figure}[htbp]
\centering
\begin{subfigure}{0.48\textwidth}
    \centering
    \includegraphics[width=\linewidth]{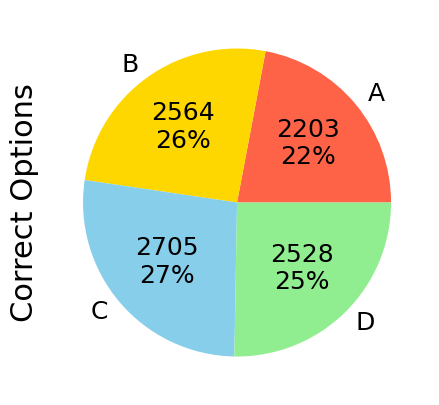} 
    \caption{Interrogative questions}
    \label{fig:interrogative}
\end{subfigure}
\hfill 
\begin{subfigure}{0.48\textwidth}
    \centering
    \includegraphics[width=\linewidth]{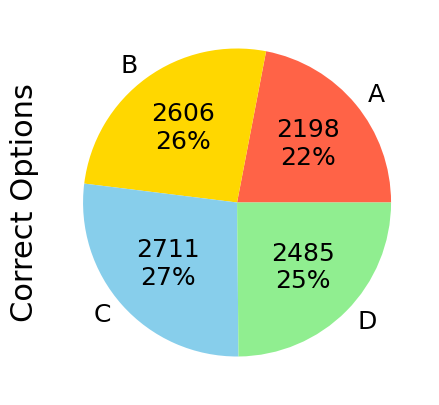} 
    \caption{Fill-in-the-blank questions}
    \label{fig:fill}
\end{subfigure}

\caption{Correct option distribution for different question types} 
\label{fig:workflow3} 
\end{figure}

We have splitted the whole dataset in into two new datasets for separating interrogative and fill-in-blank questions showing Fig. \ref{fig:workflow3}. Then, We have worked with 10 thousands question raws for each dataset. The distribution of a whole correct option into its constituent parts or categories, in other words which option occurred more frequently is represented in the following pie-charts.
These visual representation provides an insightful perspective, showing that the freqeuncy with each option came up with correct option holds close to each other. It enhances the reliability of our research work by ensuring impartiality and the absence of bias.

\chapter{Methodology, Architecture, and Model Specification}
\section{Overview of the Proposed Model}

The purpose of our proposed model is to generate relevant fill in the gaps and multiple choice questions and answers from a given article. In order to do so, we have used an open source LLM (Large language model) Llma2 (by Meta) as our base Transformer. The project was implemented using primarily the Pytorch ML framework, its libraries and dependencies. We have used the Hugging face module to import the (Llama-2-7b-hf) LLM model, which uses 7 Billion parameters for calculation. The Base LLM model was fine tuned with our collected dataset to perform better with our AQAG process. This aims to create more relevant outputs according to our Question and Answer generation use case. Before that, we collected our dataset  from English tests given to students in middle and high schools in China and preprocessed our collected data for data pipelining procedures. Then we imported all the necessary libraries needed for our AQAG process. We used prompt engineering in order to obtain desired and precise outcomes from our model. Then we used parameter quantization and parameter tokenization which makes the data easier and efficient to analyze and process. After that, creating a model instance, we fine tuned the Base LLM model with our dataset defining the training arguments. Finally, after the generation of question and answer, we evaluated the performance of our model using different evaluation metrics. The following figure represents the workflow and data-pipelining process of our system.

\section{Experimental Setup}
The proposed Question and answer generation method and other implementations are developed on Google Colab Pro, Google's in-house cloud solution for remote computing. To implement our model and all the work related to our research were done by using SSD which is 512GB, AMD Ryzen 5 5600 system with 16GB RAM and 12 GB graphics memory which is Nvidia RTX3060 .  All the work we have done is in the Python language. Also, we have used  many python libraries. For instance, json,re, pprint, pandas, torch, datasets, huggingface\_hub and transformers, scikit-learn, spacy and TensorFlow libraries. Lastly,  to produce the result Matplotlib and Seaborn libraries of python were used.

\section{Model Specification}
Llama 2 (Open source Large Language Model by Meta AI): Essentially, it is a large open source language model. This model has a solid understanding of natural language processing due to its adaptive and robust tools. To exhibit this understanding, a large amount of data is being trained to make the model adaptable for generating contextual meaningful questions and responses. Large language models are capable of achieving state-of-the-art performance in activities like question-answering, depending on the unique use case and fine-tuning.

In Machine Learning, fine-tuning refers to the process of modifying any pre-trained model by adjusting weights and parameters to enhance the performance to get the desired outcomes. This fine-tuning is a significant aspect of boosting the model’s capabilities. This whole process entails training the model on a specific dataset, allowing it to adapt its understanding and modifying the model’s previous knowledge or refining it to improve its performance in the particular task. Based on this,  the model further trains with a focus on tasks such as text summarization, language translation and question answering. Because of this training process the model improves its adaptive capability of understanding which ultimately makes it more efficient and effective in handling tasks related to summarizing text, translating languages, and providing relevant answers to questions. Essentially, fine-tuning tailors the model's skills to meet the specific requirements of these tasks, optimizing its performance for practical applications. 

So, Llama-2 can be adapted for different uses by adjusting to specific datasets during fine-tuning which make it well suited for a wide range of real world scenarios. Eventually , The objective of fine-tuning helps to provide very efficient and contextually rich language processing over many applications. Considering our desire, we have used the Llama-2's fine-tuned capabilities to make it a powerful tool in the realm of natural language understanding and automatic question and answer generation.

In our research, we have used Llama-2 with 7 billion parameters which is almost 13.5 GB.  Normally, Llama-2 comes with three different sizes and those are 7 billion, 13 billion and 70 billion. We have chosen  Llama-2 with 7 billion parameters because of the limitation of the resources. These models are very advanced and need particularly large storage so we have to work with  Llama-2 with 7 billion parameters, as it takes comparatively  less storage space than the other versions.
\begin{figure}
    \centering
    \includegraphics[width=0.5\linewidth]{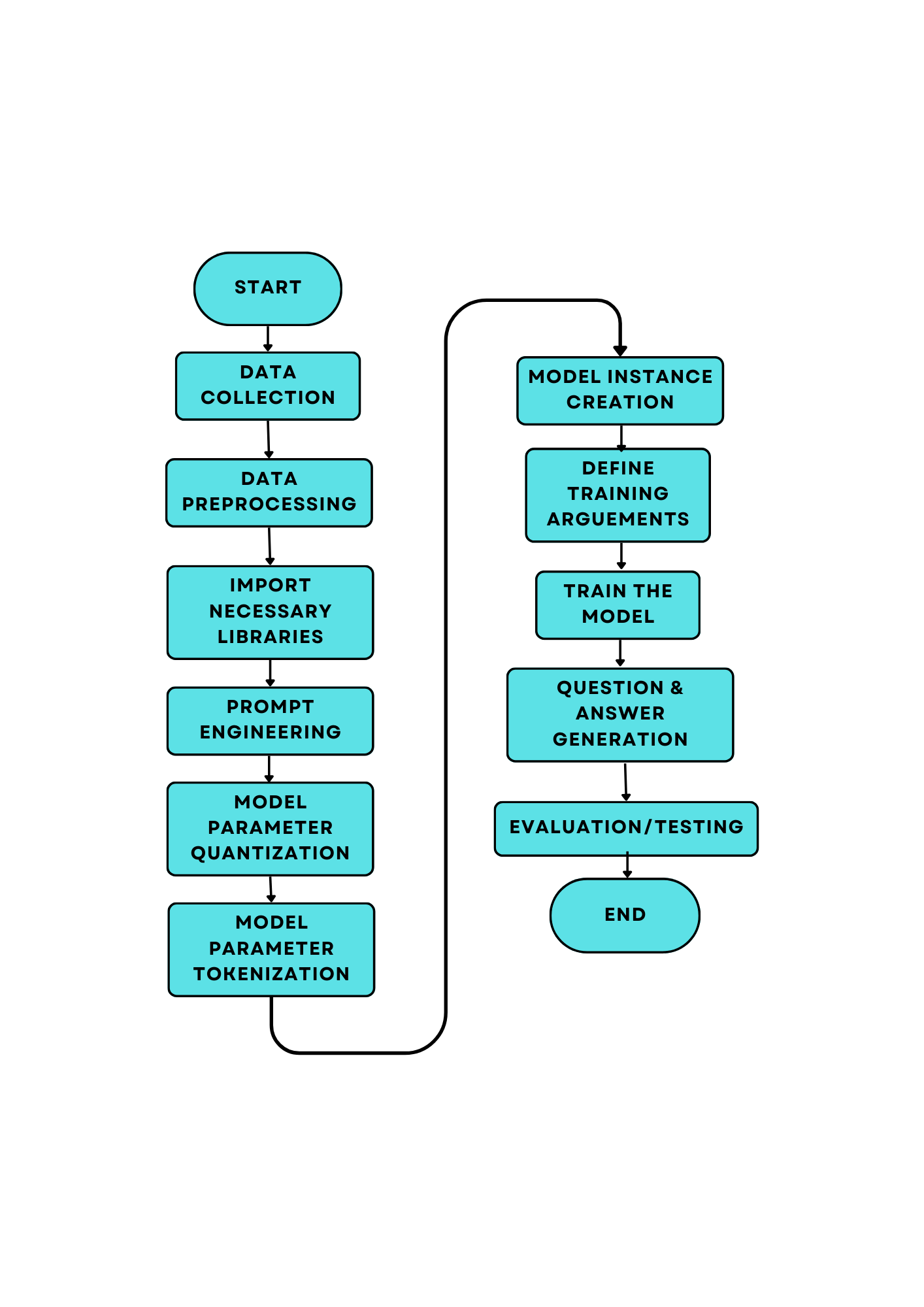}
    \caption{Data Pipeline of the proposed Model}
    \label{fig:enter-label}
\end{figure}

\section{Import Necessary Pre-requisites}
Early on in the implementation of our model, necessary environments and libraries need to be set up for constructing a robust and efficient configuration. Based on this, we installed some significant versions of python libraries, allowing us to work on solely for our desired outcomes.  For example json,re, pprint, pandas, torch,datasets,huggingface\_hub and transformers. These libraries are important for Natural Language Processing tasks.

The description and purpose of the each libraries are:
\begin{itemize}
    \item \textbf{JSON (JavaScript Object Notation):} it is a lightweight data interchange format which makes it easier for humans so that they can easily read and write. Also, easy for the machine to parse and generate. In the Natural Language process, JavaScript Object Notation (JSON) is mainly used to store and exchange structured data. It is also used to represent other elements like dataset configuration,annotations, model outputs and so on.  

     \item \textbf{Huggingface\_hub and transformers:} It provides a repository for pre-trained models and tools for working with them. Also, it is used in Question-answering; it means the process of taking the answer out of the context, producing summaries from the lengthy texts, it is helpful in text classification and creating a new text using text-generation models like GPT. We have used the Hugging Face ecosystem with the transformers library so that we are able to work with the pre-trained language model, in our case we use llama-2. 
    \item \textbf{Torch(PyTorch):} PyTorch is a comprehensive framework and an open-source deep learning library.It is written in python so it is comparatively easy to learn and use. It is mainly used for building and training models. It provides dynamic computational graphs, making it flexible and suitable for tasks like text classification, named entity recognition, image recognizing, and understanding language. 
    \item \textbf{Pandas:}  This is  the most powerful data manipulation and analysis library for Python. It is helpful for manipulating the structured data. It also explores and analyzes the data. Moreover, data is cleaned and loaded through the use of this. So it is very powerful, efficient and easy to use. 
    \item \textbf{Datasets:} We used the Datasets Library to handle our dataset.
    \item \textbf{pprint:} This is useful in  printing the data structures. It is important when it comes to debugging the code and making it easier to understand  what is happening in the code and easy to work with.
    \item \textbf{Regular Expressions:} This library determines if a particular string matches a specific pattern or not using the match function. It is mainly a strong tool that helps us to work with the regular expressions. 
    \item \textbf{LoRA:} LoRA involves the immobilization of pre-trained model weights while introducing trainable rank decomposition matrices into each layer of the Transformer architecture, resulting in a significant reduction in the number of trainable parameters available for subsequent tasks.
    \item \textbf{BitsandBytes:} It is a simple tool that wraps around CUDA functions which is designed for matrix multiplication and quantization.
\end{itemize}

\section{Prompt Engineering}
Prompt engineering is the process of developing and enhancing prompts or queries that generate certain outputs from AI models. It plays a vital role in formulating queries that assist generative AI models comprehend the intentions and complexity behind the query efficiently, not only the language. In order to obtain desired and precise outcomes from AI models, prompts are used to instruct and make adaptations to the intended behaviors. A high-grade, intelligent prompt responds appropriately by understanding user intent, generating more relevant and consistent results that positively affects the AI generated output quality. It entails the deliberate design of prompts to reduce biases, boost accuracy, and direct models towards desired outputs. Well-designed prompts resolve ambiguity, accommodate a wide range of applications, and facilitate the incorporation of subject-specific information.

The key elements of a prompt are-
\begin{itemize}
    \item \textbf{Instruction:} This is the main directive of the prompt. It instructs the model to gain the adaptability for a specific task that the model needs to perform. 

     \item \textbf{Context:} Context adds details to assist in the model's perception of the broader scene or background. It is basically additional information or more context that might help the model generate better results. 
    \item \textbf{Input Data:} This is the particular data or  information  the model needs to process. 
    \item \textbf{Output Indicator:} This component instructs the model on the intended answer type or format.
\end{itemize}
In our approach for generating multiple choice questions, We implemented two distinct prompts in our system, one is for filling in the gaps type questions and another for open-ended questions. We initialized ‘Default System Prompt’ as the instructions of the model to familiarize about the intent. Here, we provided a sample article and several sample questions from that article for the convenience of the model to help with the comprehension. The motive behind that is to make understandable the significance of questions in that particular article.While providing sample questions, we chose two types of questions to keep the diversity. In one prompt, we chose several fill in the blanks questions and in another one, we provided sample open-ended questions. This two-prompts method helps the model to grasp the ability in purpose of analyzing the significance of those questions in the article. For better learning, we incorporated the system prompt with the dataset by combining each article, corresponding question, provide options and answer. This mapping ensures the recognition and prioritization of the important keywords from each article, which thereby makes it better for generating relevant and significant questions. Overall, this methodological approach assists the model with clear instructions with necessary tools to generate insightful questions from articles.
Here is a visual representation of our prompt engineering process for generating questions and answers. We provided the model a sample article and some sample questions along with our prompt. Then after fine-tuning, the model generated some multiple choice questions.

\begin{figure}[h]
\centering 
\includegraphics[width=1.1\linewidth, height=18cm]{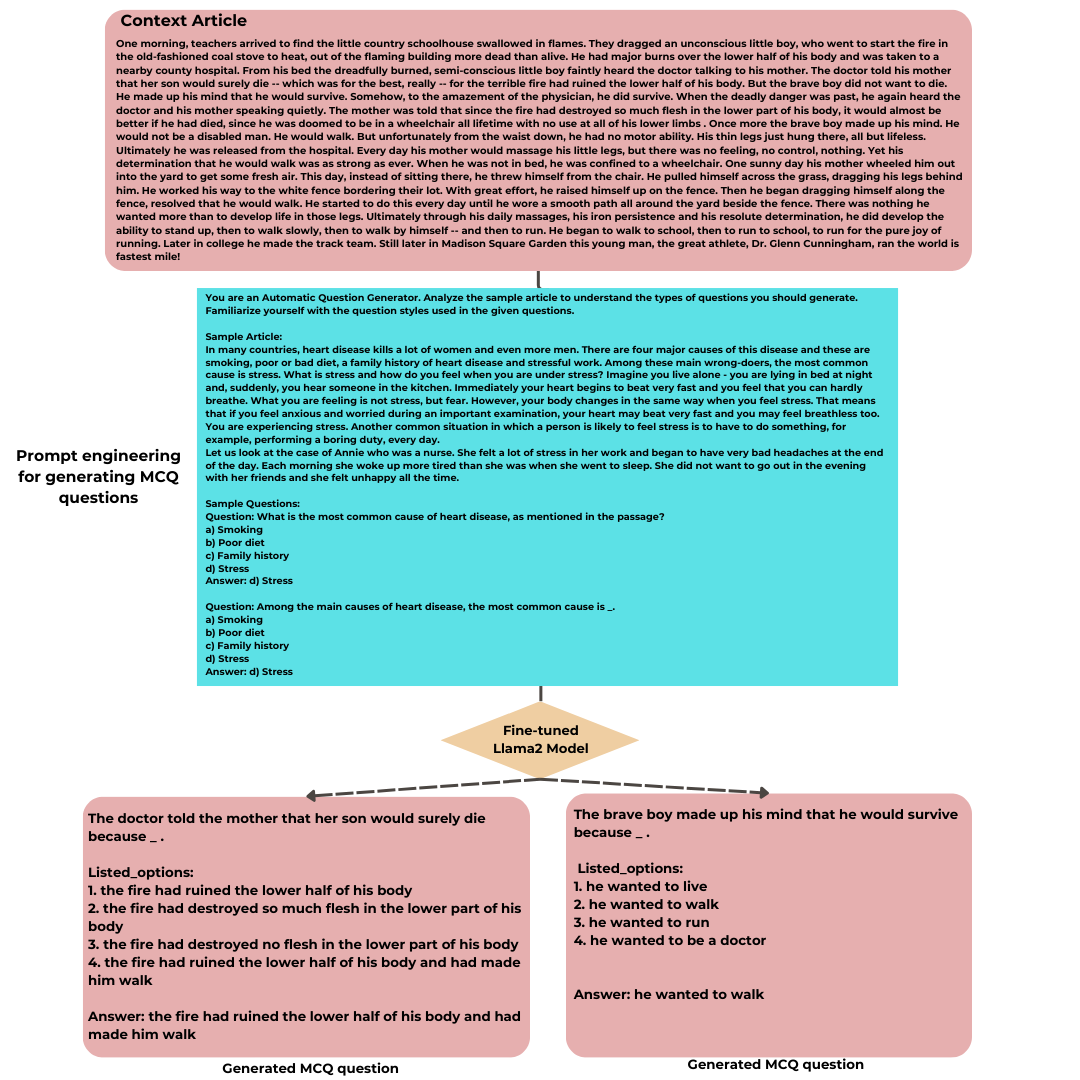}
\caption{Visualization of Prompt Engineering for our process}
\label{fig:workflow7}
\end{figure}

\section{Quantization}
In the context of the Large Language Model(LLM), quantization basically denotes the conversion of the model’s weights from higher precision to lower precision. The main objective of this is to reduce the computational and memory costs of running inference by representing the weights and activations with low-precision data types like k-bit integer (intk) instead of the usual 32-bit floating point (float32). Instead of using more complex precise numbers it simplifies those to fewer bits which helps the model to become smaller and faster so that the particular model can run on the devices which have limited resources. 

In our proposed model, we applied a 4-bit quantization technique named Normal Floating-4(NF4), a parameter of the “BitsAndBytesConfig” library that we imported from transformers. It’s one of the most efficient techniques introduced by QLoRA to minimize the memory allocations by keeping intact the performance. We enabled the 4-bit quantization model by the variable load\_in\_4bit ‘True’. For computational data type, we chose ‘float16’ as it ensures memory efficiency.

Overall, it simplified 32 bit floating point numbers into 4 bits. By doing this, we reduced the memory allocation by 8 times less. So quantization is like simplifying and rounding numbers to make them easier to work with, saving space and speeding up the process in the machine learning models. For our model llama-2 7b we used  quantization instead of 8-bit quantization. Because if we use 8 bit quantization it will need 6.52 GB of memory for execution. Allocating such a significant amount of storage solely for quantization would limit our ability to perform additional tasks, as our VRAM is already occupied with other data during that period. So instead of using 8 bit quantization we had to use 4-bit quantization because it took less amount of storage space and we can perform other tasks at that time.

To calculate how much memory can be saved by 4-bit quantization:
\begin{center}
\begin{math}
memory\_saved = [1-(bit4\_total\_memory/float32 total memory)]*100%
\end{math}
\end{center}
\section{Tokenization}
Tokenization is the process of breaking down a piece of text into smaller chunks. These smaller chunks are referred to as a token. This token is helpful because for this it is easy for the machine to process and understand the sentiment of a word. It is a basic task in Natural Language Process because it helps in language modeling, examining the sentiment and machine translation. This technique gives us a structured and organized way to represent and analyze  sentences, it helps algorithms to recognize the individuals or the subwords to understand their sentiment. Tokenization approaches boost the performance of a machine to understand languages more effectively and efficiently. In our model we created a tokenizer and our tokenizer was associated with the pretrained model.  We have set the padding token of the tokenizer to be the end of the sequence token and the padding side is configured to be on the right. Because to make the sentence equally long it needs padding like adding spaces at the end of the sentence to make them equal so it does its job using “end of sequence”. These things are needed when we try to teach our model how to handle text.

\section{Parameter Fine-Tuning}
We used SFT Trainer as our trainer which is a specifically designed model that is designed to train large language models effectively while paying close attention to memory economy and fine-tuning strategies. To maximize resource use, a gradient accumulation of 8 is paired with a batch size of 2. The Paged AdamW optimizer and chosen gradient clipping value of 0.3 provide stability, while mixed-precision training with FP16 is supported for quicker calculation. Processing text input is limited by the selected maximum sequence length of 1024.After two training epochs, the model is evaluated at intervals of 175 steps using a cosine annealing learning rate schedule. TensorBoard displays the training progress and saves checkpoints at the end of each epoch. The purpose of these hyperparameter selections and approaches is to maximize training effectiveness and performance for our selected model architecture. Then, we set a collate function to provide a consistent method for preprocessing textual data while our machine learning model was being trained. Utilizing the tokenizer of the Hugging Face Transformers library with parameters set for truncation and padding, the collate function ensures that input sequences are properly handled before they're included into the training pipeline. Finally, we trained our model using the SFT Trainer. 

The completely trained model could not be saved because of the large size and abundance of parameters in Llama2 7B causing the storage constraint to surpass. Furthermore, storing the complete model needs a large amount of memory, using up more than our storage capacity. We used 4 bit quantization and by compressing model weights and activations to 4 bits, 4-bit quantization greatly minimizes memory footprint and speeds up computations. But saving pretrained models are not yet able to fully preserve the quantized model structure. For this reason, rather than storing the complete model, we stored just the state dictionary. The learnt parameters as in weights and biases are stored in the state dictionary, which is typically enough to load and continue training or to make predictions. We used PyTorch's torch.save() function to save the state dictionary of the model. Then we used the Hugging Face Transformers library to save the configuration of the model.

In order to load our saved pretrained model, we need to follow the reversed steps of saving the model. First, we have to create an instance of the model with the saved configuration and then load the stored weights with that configured model. But while loading the model, due to the massive size of the model, the session crashes after using all available RAM before completing the loading process.

\section{Model Testing}
In the testing phase, we used the same default system prompt that we used while training along with the testing dataset articles for generating the prompt. Then we tokenized the prompt and converted the tokens to pytorch tensors. Using our fine tuned model, a tokenizer and the generated prompt, we generated the questions. Using the max new tokens and temperature parameters respectively control the created text's length and variance.

Here are some sample generated questions and answers from a sample article.

\begin{figure}[h]
\centering 
\includegraphics[width=1.05\linewidth, height=7cm]{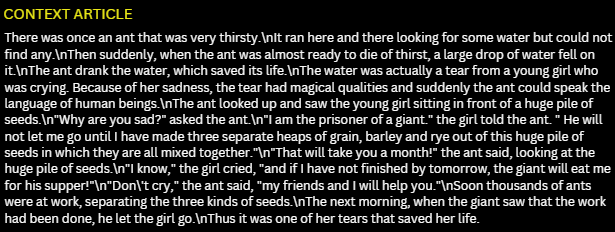}
\caption{Sample Article}
\label{fig:workflow6}
\end{figure}

{\frenchspacing After fine tuning the Llama-2 7B model with our training dataset, the model generated some insightful multiple choice questions based on our test dataset articles in the testing phase. The following image displays some questions generated from that one particular article. After generating the question there are four options and one correct answer for each question. Moreover, there is a brief explanation for each of the correct answers.}

\break
\begin{figure} [h]
\centering 
\includegraphics[width=1.0\linewidth, height=17cm]{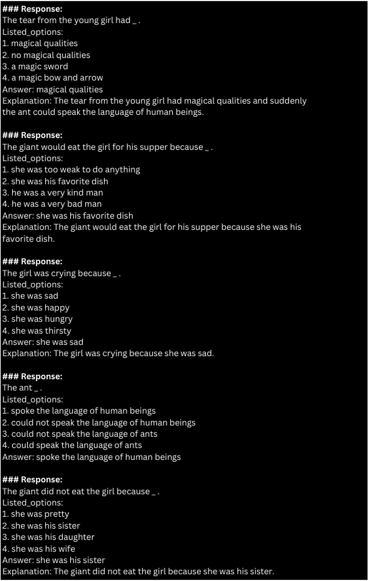}
\caption{Generated Questions}
\label{fig:workflow5}
\end{figure}
\break

\chapter{Result Analysis}
\section{Evaluation Findings}
Through the exploration of BERT-based extractive text summarization techniques in the preliminary stages of our research, implementing sentence-based multi-layered embeddings and tokenization procedures to generate comprehensive summarized texts from given context articles. Transformer-based encoding strategies proved to be good enough for articles assigned with smaller context lengths but struggled with reliable results in larger text generation tasks. Exploring the T5 model, after analyzing 150 rows of test data, about 40\% of them generated relevant readable output with a training loss of 1.5946 and a validation loss of 1.6487. This may primarily be caused by lowered training parameters due to hardware restrictions.  We selected 50 suitable outputs and evaluated their perplexity score and Semantic similarity score with the context article and test question input in 2 sets. Perplexity serves as a statistical gauge of a language model's prediction confidence for a given text sample. It quantifies the level of unexpectedness or surprises the model experiences when encountering new data. With an average perplexity score range between (197.31 - 312.48). Which is not ideal for our transformer model. Moreover, the semantic score that we found from testing was on average equal to or greater than 0.5. Both are not ideal for Large language transformer models. While BERT-based transformer models worked with the sentence-level summarization process, we struggled to implement comprehensive methods of generating questions and answers similar to the concise summarized text. 
\begin{figure}[h]
\centering 
\includegraphics[width=1.01\linewidth, height=5.25cm]{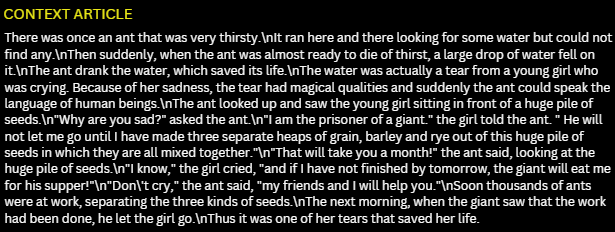}
\caption{Sample Article}
\label{fig: Context Article}
\end{figure} \break
\begin{figure}[h]
\centering 
\includegraphics[width=1.01\linewidth, height=6cm]{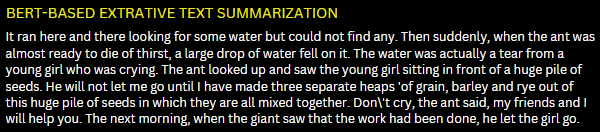}
\caption{BERT-based Extractive Text Summarization}
\label{fig: BERT-based Extractive Text Summarization}
\end{figure}

In response to challenges encountered, we transitioned our approach to leverage existing open-source Large Language Models (LLMs) for our research outcome and analysis. Our research sphere primarily focused on utilizing Meta’s Llama-2 open-sourced model. As the new wave of commercially available LLMs automatically inherits text-summarization capabilities, we aimed to utilize its features by fine-tuning with the RACE reading comprehension dataset, hoping to feed for data with our final objective of question and answer generation, improving its customized adaptability. The basic model was trained using a subset of 10,000 rows from the dataset that was particularly designated for further model fine-tuning. After numerous phases of prompt engineering tests and relevant results adhering to the intended prompt structure of the Llama-2 model using  $$<<SYS>>-<</SYS>>, [INST]/-[/INST].$$ \frenchspacing We instructed the LLM to first familiarize itself with our intended Multiple-choice question and answer structure. Finally testing the fine-tuned trainer model on unseen context articles. 

\begin{figure}[h]
\centering 
\includegraphics[width=1.0\linewidth, height=9cm]{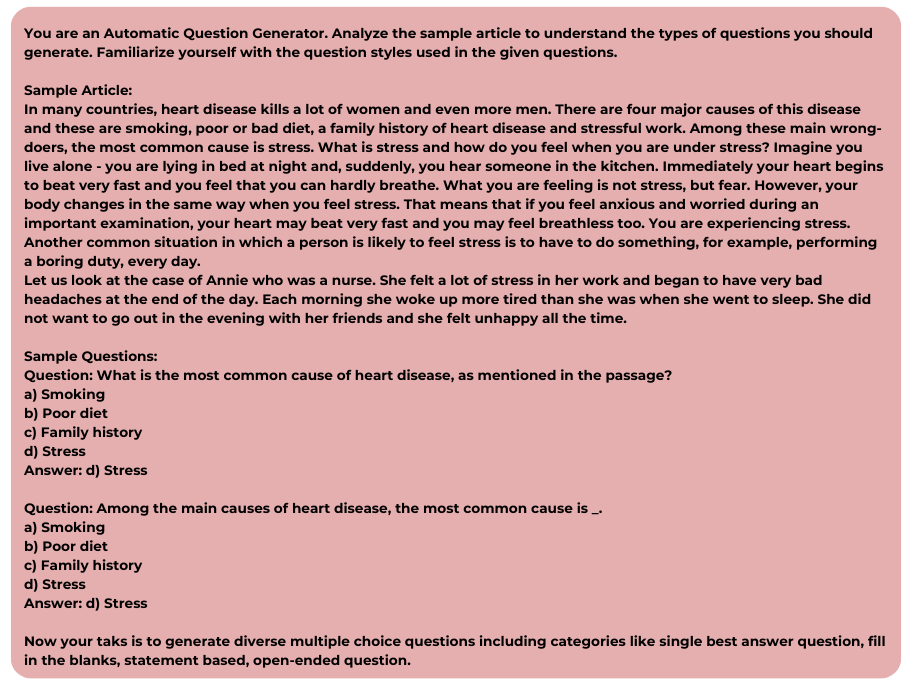}
\caption{Default System Prompt}
\label{fig: BERT-based Extractive Text Summarization}
\end{figure}
\break

\section{Evaluation Metrics}
The scientific community is striving towards finding better metrics for subsequently evaluating Question and Answer generation systems. However, there has always been a critique about using n-gram-based similarity metrics such as (ROUGE, BLEU, and METEOR) scores for evaluating the performance of Automatic Question and Answer Generation (AQAG) systems.
\\
As they generally focus on text summarization, translation, semantic relatedness and similarity index with reference inputs. The extent of semantic similarity or proximity between two or more words, sentences, or concepts is known as semantic relatedness. It illustrates how much language and communication these components have in common in terms of concepts, contexts, or associations. Low semantic relatedness denotes a weaker or less prominent relationship, whereas high semantic relatedness reflects a significant connection in meaning. In natural language processing, information retrieval, and other linguistic applications where knowing the connections between words and their meanings is critical, this idea of semantic relatedness is significant.

In our research, we worked by benchmarking our fine-tuned Llama-2 model as a whole with model perplexity scores. Perplexity is denoted as a statistical measurement that measures the level of confidence when it comes to predicting a text sample. It mainly measures how a model reacts when new data is introduced. In addition to its applications, the perplexity metric is used to compare various language models, find issues, and also help models in fine-tuning the parameters. In natural language processing, It is a very fast and efficient metric that evaluates the model performance by measuring the dataset log-likelihood. Moreover, it helps fine-tune the hyperparameters of natural language processing models on huge datasets. It is also helpful to identify if the model is overfitting or underfitting. To sum up, with the decrease in perplexity, the model produces better predictions of the text. We primarily used the WikiText-2 test set for our language model testing dataset, This contains a collection of over 100 million tokens of Wikipedia articles, which is available under the Creative Commons Attribution-ShareAlike License.
The formula for finding perplexity score,

\begin{equation}
    \text{Perplexity}(X) = \exp\left\{-\frac{1}{t}\sum_{i=1}^{t}\log P_\theta \,(\mathbf{x}_i| \mathbf{x}_{<i})\right\}
\end{equation}

After evaluation, our custom fine-tuned model on average showed similar results according to current industry benchmarks, keeping in mind the 4-bit quantization process. Our customized model shows an overall score of 6.43.\\

\begin{table}[!h]
\begin{center}
\begin{tabular}{||c c c||} 
 \hline
 Model/Quantization & q4 & q8  \\ [0.5ex] 
 \hline\hline
 llama-7b & 6.0915 & 5.9063\\ 
 \hline
 llama-13b & 5.3608 & 5.2547\\
 \hline
 llama-30b & - & -\\
 \hline
 llama-30b & - & -\\
 \hline
 llama-2-7b & 6.0398 & 5.7897\\
 \hline
 llama-2-7b-chat & 7.7853 & 7.5014\\
 \hline
 llama-2-13b & 5.2115 & 5.1005\\
 \hline
 llama-2-13b-chat & 6.7059 & 6.5361 \\ 
 \hline\hline
 llama-2-7b Fine-Tuned & 6.43 & - \\
 \hline

\end{tabular}
\caption{\label{demo-table}Perplexity on wikitext-2 test set}
\end{center}
\end{table}
For analyzing the relevance score of each question with the context, we have implemented the tfidf\_vectorizer, a feature of scikit\-learn library. By using TF\-IDF vectorizer we have converted the questions and articles in corresponding vectorized form where each word carries a vector weight. After that, we have applied cosine similarity to the vectorized article and question to find how relevant the generated question to the corresponding article. We have calculated the score on our generated questions that has been presented before. In the table, higher relevance score represents that the question is more relevant to the sample article. 
\\
\begin{table}[!h]
\begin{center}
\begin{tabular}{||c c ||} 
 \hline
 Generated Questions & Relevance Score \\ [0.5ex] 
 \hline\hline
 Question-1 & 0.55533549 \\ 
 \hline
 Question-2 & 0.51964122 \\
 \hline
 Question-3 & 0.4624515 \\
 \hline
 Question-4 & 0.75997221 \\
 \hline
\end{tabular}
\caption{\label{demo-table}Relevance Score with the context}
\end{center}
\end{table}
\pagebreak \\\\
Furthermore, for a decent multiple-choice question, it’s significantly important to set all the available options to similar types of category. Thus, the question holds a standard quality. It is crucial because it lessens biases and guarantees the assessment's fairness. So, we calculated the cosine similarity of each option with the correct answer. For that, we used the en-core-web-ig model of spacy which can give us the similarity score. In the following table, the similarity score of each option with the correct answer is illustrated. Here, a higher score indicates that the option has more similarity with the correct answer. 
\\
\begin{table}[!h]
\begin{center}
\begin{tabular}{||c c c c c||} 
 \hline
 Generated Questions & Options 1 & Options 2 & Options 3 & Options 4 \\ [0.5ex] 
 \hline\hline
 Question-1 & 1.0000001 & 0.58418226 & 0.48710752 & 0.5534273 \\ 
 \hline
 Question-2 & 0.5847328 & 1.0000001 & 0.78906226 & 0.7800066 \\
 \hline
 Question-3 & 1.0 & 0.93975097 & 0.9534698 & 0.9415601 \\
 \hline
 Question-4 & 1.0 & 0.93176795 & 0.885403561 & 0.92620562 \\
 \hline

\end{tabular}
\caption{\label{demo-table}Similarity Score of Listed Options with Correct Answer using Spacy}
\end{center}
\end{table}
\\
For evaluating both the scores, we have used the formula of cosine similarity, 
\begin{equation}
    \text{cosine\_similarity}(\mathbf{A}, \mathbf{B}) = \frac{\mathbf{A} \cdot \mathbf{B}}{\|\mathbf{A}\| \cdot \|\mathbf{B}\|}
\end{equation}
\\
\pagebreak

\section{Training Loss}
In the following, the visual representation of the decrease in training loss is illustrated, proportionately indicated improved performance over time. 


\begin{figure}[h]
\centering 
\includegraphics[width=1.1\linewidth, height=7.5cm]{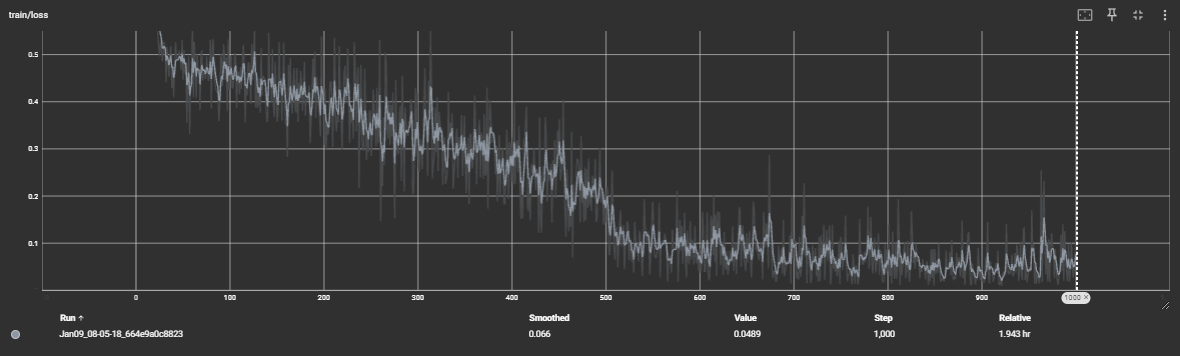}
\caption{Training Loss}
\label{fig: BERT-based Extractive Text Summarization}
\end{figure}
\break

\chapter{Conclusion}
In this paper, we talked about the importance of question and answer generation in the education sector. We propose an approach to generate questions and corresponding answers from a document using a generative large language model. It will utilize prompt engineering for generating insightful questions and corresponding answers in different formats. We believe that our work will create a new dimension to various sectors including education, e-commerce, medical and so on in this age of automation. It will be highly beneficial to both learners and educators. Besides, it will generate questions for the job applicants in terms of their qualifications and interests. It will facilitate things for the interviewers. In addition, we believe that our study will be a great assistance to the new researchers who will upgrade and add new features to this kind of approach.In future, we aim to upgrade our model which will be able to generate analytical questions and their corresponding answers. We'll improve its usability and efficiency so that it can automatically respond to the inquiries of the e-commerce customers promptly. 
  
\section{Challenges}
While progressively navigating our research journey, some challenging hurdles have emerged. We selected  Meta-Llama 2 as our base model, which had its open course license released in July 2023. Working with newly released models involves constraints in finding necessary and relevant information resources, difficulties in orchestrating the experiments, and ultimately putting in a challenging endeavor to get the desired outcomes. Moreover, owing to limitations in high-capacity RAM requirements for loading the GenAI models. We were constrained to only using the smaller 7 Billion parameter Llama 2 model. In full precision which requires 28 GB VRAM. Due to the unavailability of adequate hardware, a 4-bit quantized version was introduced to run in the local training setup. We have faced issues while training and testing our pre-trained model, sacrificing performance for model execution.  proportionately affecting the proposed model's efficiency and precision metrics. Talking about one crucial factor in facing these challenges, is insufficient funding for access to better GPU hardware and financial scarcity for undergraduate students. Which significantly impedes our planned research endeavor. 

\section{Limitations}
While implementing the Automatic Question and Answer Generation model, we have faced some constraints. First of all, as we know, for fine-tuning a pre-trained model on a certain task, it should not encounter any biases during the training phase. But, if we carefully look at our generated factual questions in testing, the majority of the questions are like “What is the main idea of the passage?”. The suspected reason lies in our training dataset, which carries this type of question excessively in most articles, resulting in the model bias to this type of conceptual question. Furthermore, we have encountered issues while generating questions when the run-time gets larger. Secondly,  the automatic question-answer generation system can generate multiple questions simultaneously, but the overall quality is inconsistent, particularly those questions that are generated later on in the testing phase. This is because of the run-time limitations of V-Ram(higher specification of GPU). Moreover, This approach could not produce analytical questions and their answers as we only focus on the MCQ-based questions. The system can generate questions and their respective answers based on a provided textual article, but it lacks the ability to perform this task directly from a PDF or document format. Furthermore, our proposed model has been fine-tuned focusing only on the English language, therefore it can’t handle different languages which limits the potential of the model.  

\section{Future Work}
In future work, we aim to extend upon our proposed model methodology and look into the exploration of newer Large Language Models (LLMs) with better performance emerging in the dynamic landscape of generative language models. Given the continuous expansion in the realm of work in generative AI (GenAI), our research endeavors will be dedicated to investigating, incorporating, and benchmarking against the latest advancements in the industry. Furthermore, our experimental study exhibits a relative bias toward generating conceptual questions based on a given test article and a decline in question generation quality over multiple test phases. We plan to investigate this using downstream layer representation strategies employing Explainable AI (XAI) techniques. Future studies also plan to expand our research agenda to include the multilingual and multi-file document (MfD) model interpretability, striving for better performance, inclusivity, and efficacy.

\phantomsection

\addcontentsline{toc}{chapter}{Bibliography}



\end{document}